%% file: neurips_2025.tex
\documentclass{article}

\PassOptionsToPackage{numbers, compress}{natbib}
\usepackage[dblblindworkshop, final]{neurips_2025}

\usepackage[utf8]{inputenc} 
\usepackage[T1]{fontenc}    
\usepackage{hyperref}       
\usepackage{url}            
\usepackage{booktabs}       
\usepackage{amsfonts}       
\usepackage{nicefrac}       
\usepackage{microtype}      
\usepackage{xcolor}         
\usepackage[most]{tcolorbox}
\usepackage{wrapfig}

\input{commands/mathcommand}

\input{commands/mycommand}

\newcommand{\myalg}{\code{\textbf{TSSS}}\xspace}
\newcommand{\mytitle}{Think Straight, Stop Smart: Structured Reasoning \\for Efficient Multi-Hop RAG}
\title{\mytitle}
\workshoptitle{Think Straight, Stop Smart: Structured Reasoning for Efficient Multi-Hop RAG}

\author{Jihwan Bang\hspace{2em}Juntae Lee\hspace{2em}Seunghan Yang\hspace{2em}Sungha Choi\thanks{indicates the corresponding author.} \\
{Qualcomm AI Research\thanks{Qualcomm AI Research is an initiative of Qualcomm Technologies, Inc.}, Qualcomm Korea YH, Seoul, Republic of Korea} \\ 
{\texttt {\small\{jihwbang, juntlee, seunghan, sunghac\}@qti.qualcomm.com}}}

\begin{document}

\maketitle


\begin{abstract}
Multi-hop retrieval-augmented generation (RAG) is a promising strategy for complex reasoning, yet existing iterative prompting approaches remain inefficient. They often regenerate predictable token sequences at every step and rely on stochastic stopping, leading to excessive token usage and unstable termination. We propose \myalg (\underline{\textbf{T}}hink \underline{\textbf{S}}traight, \underline{\textbf{S}}top \underline{\textbf{S}}mart), a structured multi-hop RAG framework designed for efficiency. \myalg introduces (i) a template-based reasoning that caches recurring prefixes and anchors sub-queries to the main question, reducing token generation cost while promoting stable reasoning, and (ii) a retriever-based terminator, which deterministically halts reasoning once additional sub-queries collapse into repetition. This separation of structured reasoning and termination control enables both faster inference and more reliable answers. On HotpotQA, 2WikiMultiHop, and MuSiQue, \myalg achieves state-of-the-art accuracy and competitive efficiency among RAG-CoT approaches, highlighting its effectiveness in efficiency-constrained scenarios such as on-device inference.
\end{abstract}

\input{contents/01_Intro}

\input{contents/02_Method}

\input{contents/03_Exp}
\input{contents/04_Related}
\input{contents/05_Conclusion}

\newpage
\bibliographystyle{unsrt}
\bibliography{neurips.bib}


\newpage
\appendix
\input{contents/06_Appendix}

\end{document}

%% file: commands/mathcommand.tex
\usepackage{amsmath,amsfonts,bm}

\def\eqref#1{equation~\ref{#1}}

\def\1{\bm{1}}

\DeclareMathAlphabet{\mathsfit}{\encodingdefault}{\sfdefault}{m}{sl}
\SetMathAlphabet{\mathsfit}{bold}{\encodingdefault}{\sfdefault}{bx}{n}



%% file: commands/mycommand.tex
\usepackage{geometry}
\usepackage{booktabs} 
\usepackage{bbm}
\usepackage{mathtools}
\usepackage{nccmath}
\usepackage{setspace}

\usepackage{caption}
\usepackage{subcaption}

\usepackage[linesnumbered,ruled,vlined]{algorithm2e}

\SetKwInput{KwInput}{Input}                
\SetKwInput{KwOutput}{Output}              

\SetCommentSty{mycommfont}

\SetAlCapSty{algcapsty}

\usepackage[T1]{fontenc}
\usepackage{wrapfig,lipsum,booktabs}

\usepackage{soul}
\usepackage{dsfont}
\usepackage{enumerate}
\usepackage{enumitem}

\usepackage{amsmath}
\usepackage{amsfonts}
\usepackage{bbm}
\usepackage{dsfont}
\usepackage[Symbol]{upgreek}
\usepackage{lscape}
\usepackage{caption}
\usepackage{balance}
\usepackage{xspace}
\usepackage{float}
\usepackage{kotex}

\usepackage{wasysym}
\usepackage{xcolor}
\usepackage{multirow}
\usepackage{array, boldline, rotating}

\usepackage{amssymb}
\usepackage{pifont}
\renewcommand*\eqref[1]{(\ref{#1})}

\newcommand{\yes}[1]{\textcolor{blue}{[YES]}}
\newcommand{\no}[1]{\textcolor{orange}{[NO]}}
\newcommand{\na}[1]{\textcolor{gray}{[N/A]}}
\newcommand{\eg}{\emph{e.g.,~}}
\newcommand{\ie}{\emph{i.e.,~}}

\newcommand{\myparagraph}[1]{\vspace{0.07cm}\noindent\textbf{#1}~}

\def\code#1{\texttt{#1}}

\definecolor{LightCyan}{rgb}{0.88,1,1}
\definecolor{Blue}{rgb}{0, 0.5, 1}
\definecolor{Green}{rgb}{0.0, 0.8, 0.0 }
\definecolor{Red}{rgb}{0.95, 0.55, 0.6}
\definecolor{Skyblue}{rgb}{0.6, 0.6, 0.95 }

\NewDocumentEnvironment{suptitle}{}{
    \onecolumn
}{}

\NewDocumentCommand{\supptitle}{s}{
\begin{suptitle}
        \centering
        \rule{\textwidth}{0.03cm}\\[0.1cm]
        -Supplementary Material-\\[0.2cm]
        {\Large 
            \textbf{\mytitle }
        }\\
        \rule{\textwidth}{0.03cm}\\[0.2cm]
\end{suptitle}}

\NewDocumentCommand{\summarytitle}{s}{
\begin{summarytitle}
        \centering
        \rule{\textwidth}{0.03cm}\\[0.1cm]
        -Summary of Changes-\\[0.2cm]
        {\Large 
            \textbf{\mytitle }
        }\\
        \rule{\textwidth}{0.03cm}\\[0.2cm]
\end{summarytitle}}

%% file: contents/01_Intro.tex
\section{Introduction}

On-device large language models (LLMs)~\cite{grattafiori2024llama, team2025gemma, abdin2024phi} are increasingly appealing for privacy-preserving and latency-sensitive applications such as personal assistants, mobile knowledge agents, and offline reasoning systems~\cite{bang2024crayon}. Unlike server-scale LLMs, however, on-device models operate with orders of magnitude fewer parameters, which severely limits their reasoning capacity~\cite{dang2025reinforcement}. At the same time, computational efficiency is critical: every generated token incurs latency and energy cost, making it essential to reduce token generation during inference~\cite{lee2025chain}.

Multi-hop question answering (QA)~\cite{yang2018HotpotQA, trivedi2022MuSiQue, xanh2020_2wikimultihop} exemplifies these challenges. It requires combining information across multiple documents, often through iterative reasoning. Multi-hop retrieval-augmented generation (RAG) has therefore been studied extensively as a powerful strategy for improving reasoning accuracy~\cite{shao2023enhancing, trivedi2023interleaving, press2023measuring, wu2025composerag, jiang2025retrieve}. However, most existing methods are designed for larger server-based models and prioritize accuracy gains over efficiency. Iterative reasoning loops frequently regenerate predictable prefixes or boilerplate sub-questions, and weak termination control often leads to duplicated queries. These inefficiencies, while tolerable in server settings, make current multi-hop RAG methods impractical for efficiency-constrained on-device inference.  

\begin{figure}[t]
    \centering
    \includegraphics[width=0.9\linewidth]{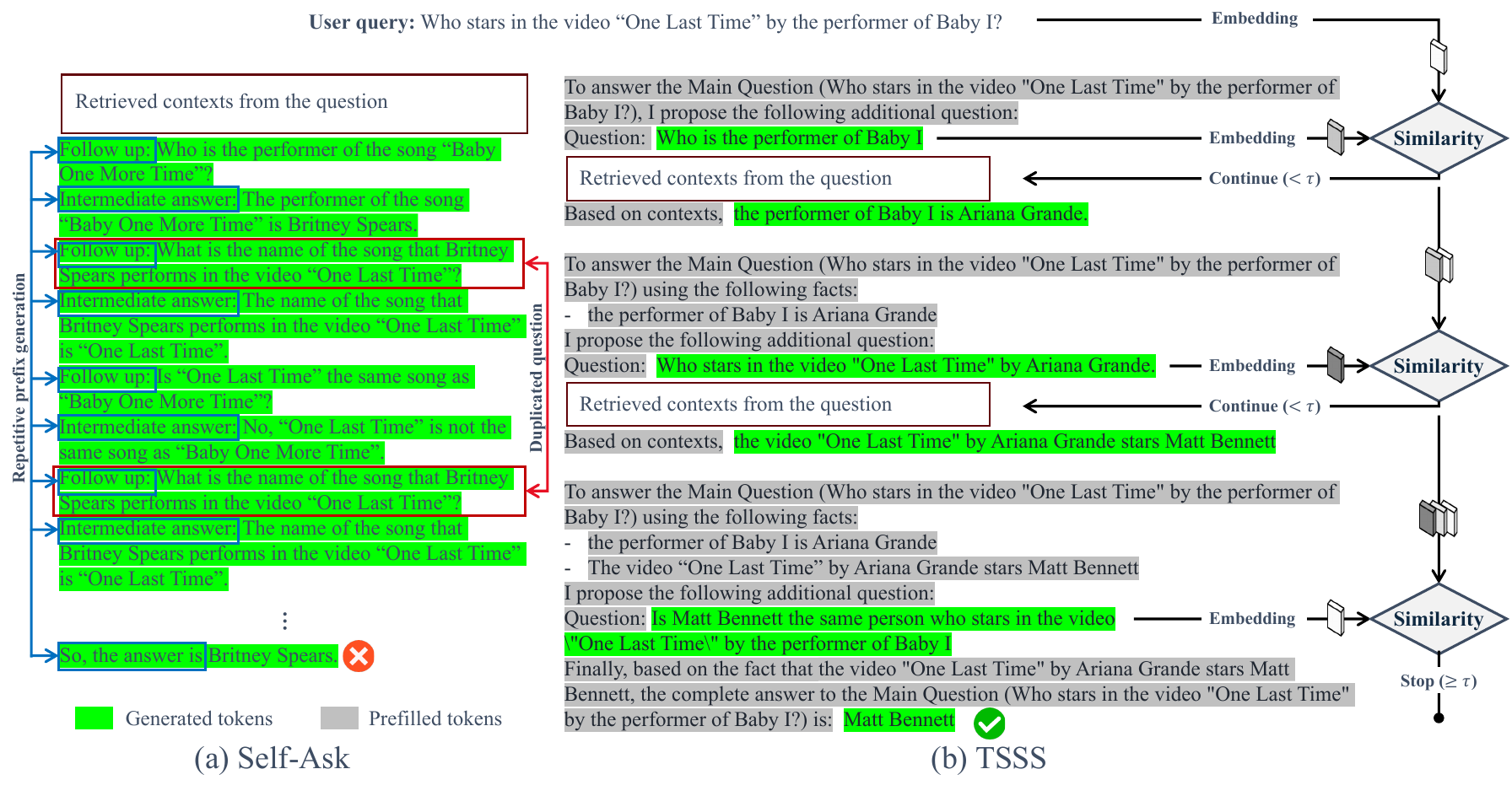}
    \vspace{-0.3cm}
    \caption{\textbf{Comparison of multi-hop reasoning processes.} 
(a) The baseline method (\eg Self-Ask) suffers from inefficiencies such as repetitive prefix generation and duplicated queries, which increase token generation usage and may lead to incorrect answers. 
(b) Our \myalg framework addresses these issues through a \textit{structured reasoning template} that reuses recurring token sequences and anchors sub-queries to the main question, together with a \textit{retriever-based terminator} that halts reasoning once further queries collapse into repetition. This combination reduces token usage while maintaining accuracy, resulting in faster and more reliable inference.}
    \label{fig:main}
    \vspace{-0.6cm}
\end{figure}

Recent efforts such as EfficientRAG~\cite{zhuang2024efficientrag} begin to address this gap by modeling when to stop reasoning. Yet, termination alone is insufficient: as shown in~\autoref{fig:main}a, token usage also balloons from repetitive and predictable generation patterns, which not only increase inference cost but may also obscure the final answer. In addition, EfficientRAG requires a dedicated termination module trained on external data. While such training can be performed on servers, it introduces additional cost, model complexity, and potential domain dependence. By contrast, Our approach requires no additional parameter training; only heuristic template design and hyperparameter selection are performed, making it effectively training-free. This plug-and-play design makes it lightweight and directly deployable to diverse on-device models without fine-tuning, while still tackling the two fundamental obstacles in efficient multi-hop RAG: \textit{predictable repetition} and \textit{generator-based termination}.

To this end, we propose \myalg (\underline{\textbf{T}}hink \underline{\textbf{S}}traight, \underline{\textbf{S}}top \underline{\textbf{S}}mart), a structured multi-hop RAG framework explicitly designed for efficient on-device reasoning. \myalg introduces two key innovations:
(i) \textit{Template-based reasoning}, which encodes recurring reasoning traces as structured templates and prefills them at each hop. This systematically eliminates redundant token generation (e.g., repeated prefixes) while keeping sub-queries anchored to the main question, thereby improving both efficiency and stability.  
(ii) \textit{Retriever-based terminator}, which shifts stopping control from the generator to the retriever. By detecting when newly generated queries collapse into repetition, the retriever deterministically halts the loop, preventing unnecessary duplication.  

Together, these innovations enable \myalg to maintain alignment with the main question while reducing token usage. As illustrated in~\autoref{fig:main}b, unlike baseline methods such as Self-Ask, which waste tokens on repetitive patterns and duplicated queries, \myalg generates concise and well-structured reasoning steps, leading to faster and more reliable inference.  

Our contributions are: (1) We formally identify predictable repetition and generator-based termination as fundamental efficiency bottlenecks in multi-hop RAG; (2) We introduce \myalg, combining a structured reasoning template with a retriever-based terminator to reduce token usage while maintaining reasoning accuracy; (3) We demonstrate that \myalg achieves substantial token savings alongside accuracy improvements, enabling practical multi-hop reasoning in on-device settings.

%% file: contents/02_Method.tex
\section{Methodology}
\subsection{Template-Based Reasoning with KV-Cache}

In standard multi-hop prompting, the LLM generates entire reasoning traces in free-form natural language. This approach is token-inefficient, since recurring prefixes such as restating the main question, retrieved evidence, or scaffolding phrases are regenerated at every iteration. Without structural constraints, the model expends computation on predictable token sequences rather than focusing on the essential reasoning content.

To address this inefficiency, we introduce a \textit{template-based reasoning framework} that leverages the KV-Cache (See~\autoref{app:kv_cache} for details). Each reasoning step is embedded into a fixed scaffold (gray regions in~\autoref{fig:main}b), where the scaffold tokens are treated as prefilled and their Key-Value states can be pre-computed and cached. As a result, the LLM only generates the variable components (green regions), such as the specific sub-query or extracted evidence. This design significantly reduces the number of newly generated tokens at each hop. By reusing the cached Key-Value states for the prefilled parts, it also lowers the computational cost for each iteration, resulting in a substantial decrease in overall latency.

Additionally, the template explicitly incorporates the main question and accumulated evidence at every iteration. By consistently anchoring sub-queries to these elements, the framework not only improves efficiency but also promotes stability in reasoning traces, mitigating the tendency of sub-queries to drift away from the original task (See~\autoref{app:template} for prompt template).

\subsection{Retriever-Based Terminator}

A common limitation of prior iterative prompting methods (\eg Self-Ask) is the generation of near-duplicate or semantically overlapping sub-queries. Such repetition increases token usage and retrieval cost without contributing new information. To prevent this inefficiency, we introduce a \textit{retriever-based termination rule} that deterministically decides when to halt the reasoning process.

Formally, let the current sub-query at iteration $i$ be denoted as $q_i$, the main question as $q_m$, and the set of previously generated sub-queries as $\{q_1, \cdots, q_{i-1}\}$. We compute embedding representations $\phi(\cdot)$ for each query from the retriever, and define a similarity score:
\begin{equation} 
    \mathrm{score}(q_i) = \max_{q \in \mathcal{Q}_i}{\cos(\phi(q_i), \phi(q))},
\end{equation}
where $\cos(\cdot, \cdot)$ denotes cosine similarity, and $\mathcal{Q}_i = \{q_m\} \cup \{q_1, \cdots, q_{i-1}\}$.

If $\mathrm{score}(q_i) \geq \tau$, the process terminates and the model produces the final answer; otherwise, reasoning continues with the next sub-query. This retriever-based termination prunes repetitive queries early, eliminating unnecessary retrievals and token generations. Consequently, it improves efficiency (fewer iterations, fewer tokens). In our experiments, we set $\tau=0.85$, which we found to balance between avoiding duplication and allowing sufficient reasoning depth (See~\autoref{app:tau} for details).

%% file: contents/03_Exp.tex
\section{Experiment}

\setlength{\tabcolsep}{4pt}
\begin{table*}[t]
  \centering
  \small
  \begin{tabular}{@{}lccccccccc@{}}
    \toprule
                  &       \multicolumn{3}{c}{HotpotQA}                            &     \multicolumn{3}{c}{2WikiMultiHop}                                  &   \multicolumn{3}{c}{MuSiQue}        \\ 
    Method        &  EM($\uparrow$) &  $ACC_L$($\uparrow$) & $T$($\downarrow$) &  EM($\uparrow$) &  $ACC_L$($\uparrow$) & $T$($\downarrow$)  &  EM($\uparrow$) &  $ACC_L$($\uparrow$) & $T$($\downarrow$)  \\
    \cmidrule(lr){1-1} \cmidrule(lr){2-4} \cmidrule(lr){5-7} \cmidrule(lr){8-10}
    No-RAG                                  &    17.7         &       29.1          &     0.1         &       16.6       &      16.6       &       0.2      &     2.6         &      7.2       &     0.2    \\  
    Standard-RAG                            &    27.7         &       40.1          &     3.3         &       11.3       &      16.2       &       3.7      &     4.1         &      7.6       &     3.3    \\  \cmidrule(lr){1-10}
    Self-Ask~\cite{press2023measuring}      &      25.6           &       37.4          &     15.2        &       21.7       &      31.4       &       13.6     &     8.2         &      14.6      &     18.0    \\
    Iter-RetGen~\cite{shao2023enhancing}    &    30.4         &       43.5          &     27.9        &       12.7       &      17.8       &       29.9     &     5.8         &      10.6      &     27.2     \\
    IRCoT~\cite{trivedi2023interleaving}    &    28.0         &       47.7          &     19.2        &       23.8       &      35.8       &       24.9     &     6.5         &      13.8      &     21.2    \\ \cmidrule(lr){1-10}
    \myalg (Ours)       &     \textbf{34.1}    &    \textbf{50.9}    &    \textbf{8.1}    &    \textbf{33.6}    &    \textbf{42.3}    &    \textbf{9.0}    &    \textbf{14.5}    &    \textbf{22.8}    &    \textbf{9.5}     \\
    \bottomrule
    
    \end{tabular}
    \caption{\textbf{Overall performance comparison of \myalg with baselines.} The evaluation metrics EM and $ACC_L$ are defined in~\autoref{app:eval_metrics}, and $T$ indicates the average inference time (seconds) per sample. \textbf{Bold} marks the best score; for $T$, it indicates the lowest among RAG-CoT methods.}
    \vspace{-0.5cm}
  \label{tab:main}
\end{table*}

\subsection{Implementation Details}
We adopt three benchmark datasets: HotpotQA~\citep{yang2018HotpotQA}, 2WikiMultiHopQA~\citep{xanh2020_2wikimultihop}, and MuSiQue~\citep{trivedi2022MuSiQue}. Wikipedia passages serve as the 21M retrieval corpus for all datasets. To verify the effectiveness of \myalg, we select several baselines ranging from No-RAG to RAG-CoT (e.g., Self-Ask~\citep{press2023measuring}, Iter-RetGen~\citep{shao2023enhancing}, and IRCoT~\citep{trivedi2023interleaving}), focusing on methods that do not require additional training cost. For fair comparison, all methods use Llama3.1-8B~\citep{grattafiori2024llama} as the generator.
To reduce retrieval cost, we use the FAISS library~\citep{douze2024faiss} with the e5-base-v2 retriever~\citep{wang2022text}, retrieving 3 documents per query. Baseline implementations are based on the open-source RAG framework FlashRAG~\citep{jin2025flashrag}. For iterative RAG-CoT methods (e.g., Iter-RetGen, IRCoT), we set the maximum number of retrieval iterations to 10 to allow self-termination. All experiments are conducted on a single NVIDIA H100 GPU.

\subsection{Results}

\myparagraph{Overall performance.}~\autoref{tab:main} summarizes the results on HotpotQA, 2WikiMultiHop, and MuSiQue. \myalg achieves the best EM and $ACC_L$ across all three benchmarks while also maintaining competitive inference efficiency. Compared to Standard-RAG, \myalg is about 2–3 times slower (e.g., 8.1s vs.\ 3.3s on HotpotQA, 9.0s vs.\ 3.7s on 2WikiMultiHop, and 9.5s vs.\ 3.3s on MuSiQue), but this overhead stems from structured iterative reasoning that substantially boosts accuracy. Importantly, \myalg is significantly faster than other RAG-CoT methods. For example, it reduces inference time by more than half compared to Iter-RetGen (e.g., 9.0s vs.\ 29.9s on 2WikiMultiHop, 9.5s vs.\ 27.2s on MuSiQue) while keeping higher performance. These results highlight its key strength: \textit{effective multi-hop reasoning with both superior accuracy and practical efficiency}.

\begin{figure}[t]
    \begin{minipage}{0.48\linewidth}
        \centering
        \includegraphics[width=\linewidth]{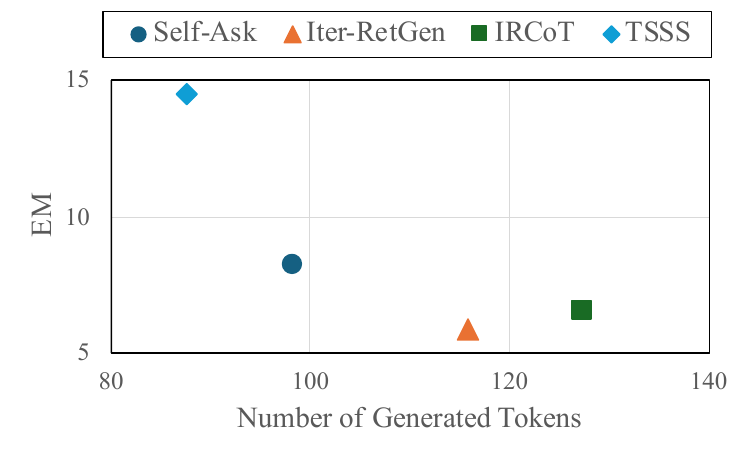}
        \vspace{-0.6cm}
        \caption{\textbf{Performance vs. efficiency on the MuSiQue dataset.} \myalg achieves a superior performance-efficiency trade-off, with the highest EM score and the fewest tokens.}
        \label{fig:perf_eff}
    \end{minipage}
    \hfill
    \begin{minipage}{0.48\linewidth}
        \centering
        \includegraphics[width=\linewidth]{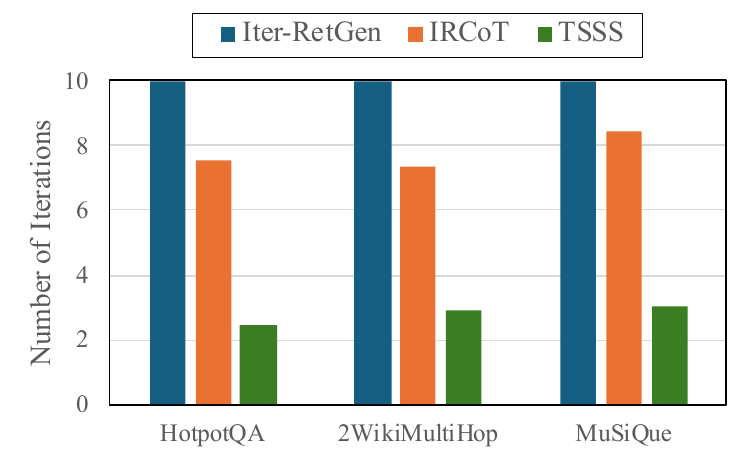}
        \vspace{-0.6cm}
        \caption{\textbf{Effect of retriever-based terminator.} \myalg proves efficient and stable termination mechanism by stopping fewer retrieval iterations than the other baselines.}
        \label{fig:n_retr}
    \end{minipage}
    \vspace{-0.5cm}
\end{figure}

\myparagraph{Performance vs. efficiency.}
\autoref{fig:perf_eff} shows the trade-off between EM performance and the number of generated tokens on the MuSiQue dataset (See~\autoref{app:ablation} for other datasets). In this setting, methods that appear closer to the upper-left corner achieve better efficiency while maintaining strong accuracy. Compared to existing baselines, \myalg achieves the highest EM score with the fewest generated tokens. This indicates that our method is not only more accurate but also significantly more efficient in terms of token usage. 


\myparagraph{Effectiveness of retriever-based terminator.}
Appropriate iteration is crucial for both performance and efficiency. As shown in~\autoref{fig:n_retr}, Iter-RetGen and IRCoT often require too many iterations, which leads to unnecessary overhead. In contrast, our method derives the answer with only 2–3 iterations, which can be regarded as a reasonable number. This effectiveness comes from the retriever-based terminator, which halts once sufficient context is gathered, enabling efficient reasoning without sacrificing accuracy. 

%% file: contents/04_Related.tex
\section{Related Works}
Retrieval-augmented generation (RAG) is a key approach for knowledge-intensive tasks but struggles with multi-hop questions.
SuRe~\cite{kimsure} proposes a framework integrating reasoning-aware retrieval and answer synthesis, while RECOMP~\cite{xu2023recomp} leverages query decomposition and retrieval fusion.
RePlug~\cite{shi2024replug} enhances single-turn RAG by incorporating structured intermediate reasoning signals during retrieval.
However, these single-turn methods often fail when intermediate reasoning is required, motivating iterative approaches (\ie RAG-CoT).
Self-Ask~\cite{press2023measuring} decomposes queries into follow-up questions, IRCoT~\cite{trivedi2023interleaving} interleaves retrieval and reasoning, and Iter-RetGen~\cite{shao2023enhancing} introduces an iterative retrieval–generation loop for more relevant evidence.

%% file: contents/05_Conclusion.tex
\section{Conclusion}
We presented \myalg, a multi-hop RAG framework that enhances efficiency with template-based reasoning and a retriever-based terminator. It improves accuracy on HotpotQA and MuSiQue, and yields faster inference on 2WikiMultiHop with minor accuracy loss, showing its value in efficiency-constrained settings (See~\autoref{app:futurework} for limitations).


%% file: contents/06_Appendix.tex
\supptitle

\section{Preliminary: KV-Cache for Efficient Decoding}
\label{app:kv_cache}

In Transformer-based LLMs, auto-regressive decoding at step $t$ requires computing the interaction between the current token and all $t$ previously generated tokens. Let $T$ be the total sequence length and $d$ the model (hidden) dimension. For clarity, the following costs are per layer and omit constant factors from the number of heads.

\paragraph{Without KV-cache.}
At decoding step $t$, a naive implementation recomputes the entire prefix of length $t$. The cost is $\mathcal{O}(t\,d^2)$ for the $Q,K,V$ projections and $\mathcal{O}(t^2 d)$ for self-attention (scores and the weighted sum). Summed over all steps, this is formulated as below. 
\[
\sum_{t=1}^{T} \big(\mathcal{O}(t\,d^2) + \mathcal{O}(t^2 d)\big)
= \mathcal{O}(T^2 d^2) + \mathcal{O}(T^3 d),
\]
i.e., cubic scaling in $T$ due to recomputing past states at every step. (By contrast, a single full-sequence forward pass of length $T$ costs $\mathcal{O}(T d^2 + T^2 d)$.)

\paragraph{With KV-cache.}
At step $t{+}1$, we keep the cached keys/values $K_{1:t}, V_{1:t}$ and only compute the incremental components $K_{t+1}, V_{t+1}$. The attention becomes
\[
\mathrm{Attn}(Q_{t+1}, K_{1:t+1}, V_{1:t+1})
= \mathrm{Attn}(Q_{t+1}, [K_{1:t} \Vert K_{t+1}], [V_{1:t} \Vert V_{t+1}]),
\]
where $\Vert$ denotes concatenation along the sequence dimension. The per-step cost is then
$\mathcal{O}(d^2)$ for the new token’s projections plus $\mathcal{O}(t\,d)$ for attending to $t$ cached tokens, i.e., $\mathcal{O}(d^2 + t\,d)$. Summed over $T$ steps, the total cost is calculated as follow.
\[
\mathcal{O}(T d^2 + T^2 d).
\]
Thus, the KV-cache removes the cubic term caused by recomputation, but total decoding time remains quadratic in $T$ because each new token attends to all prior tokens.\footnote[1]{If attention is restricted to a fixed window $W$ (e.g., sliding-window attention), the attention cost becomes $\mathcal{O}(T\cdot W \cdot d)$, which is linear in $T$ for fixed $W$.}
The memory footprint of the KV-cache is $\mathcal{O}(T\,d)$ per layer.

\section{Prompt Template for \myalg} 
\label{app:template}
The following template illustrates the structured reasoning format used in \myalg for multi-hop question answering. In this template, \{main question\} represents the original user query that the reasoning process aims to answer. Each \{LLM-generated Question $i$\} is a sub-question created by the model at iteration $i$-th to gather additional evidence. The placeholder \{Retrieved contexts with LLM-generated Question $i$\} refers to the top-k passages retrieved from the external knowledge source (\eg Wikipedia) based on that sub-question. \{LLM-generated Response $i$\} is the model’s answer to the sub-question using the retrieved contexts.

\begin{tcolorbox}[fonttitle=\small\bfseries,
fontupper=\scriptsize\sffamily,
fontlower=\fon{put},
enhanced,
left=2pt, right=2pt, top=2pt, bottom=2pt,
title=Prompt template for \myalg]
\begin{lstlisting}[]
To answer the Main Question ({main question}), I propose the following additional question:
Question: {LLM-generated Question1} 
{Retrieved contexts with LLM-generated Question1}
Based on the contexts, {LLM-generated Response1}

To answer the Main Question ({main question}), using the following facts: 
- {LLM-generated Response1}
I propose the following additional question:
Question: {LLM-generated Question2} 
{Retrieved contexts with LLM-generated Question2}
Based on the contexts, {LLM-generated Response2} 

To answer the Main Question ({main question}), using the following facts: 
- {LLM-generated Response1}
- {LLM-generated Response2}
I propose the following additional question:
Question: {LLM-generated Question3} 
{Retrieved contexts with LLM-generated Question3}
Based on the contexts, {LLM-generated Response3}
...

Finally, based on the fact that {LLM-generated ResponseN}, the complete answer to the 
main question ({main question}) is: 

\end{lstlisting}
\end{tcolorbox}

\section{Evaluation Metrics}
\label{app:eval_metrics}

During the evaluation phase, we adopt exact-match (EM) as our primary metric, which determines whether the predicted answer exactly matches the golden answer. To further refine our evaluation, we employ an LLM-as-Judge approach~\cite{zheng2023judging}, using GPT-4o~\cite{hurst2024gpt} as the evaluation model to assess whether the predicted answer is correct. This accuracy metric is referred to as $ACC_L$. The evaluation prompt is as follows.

\begin{tcolorbox}[fonttitle=\small\bfseries,
fontupper=\scriptsize\sffamily,
fontlower=\fon{put},
enhanced,
left=2pt, right=2pt, top=2pt, bottom=2pt,
title=Prompt template for LLM-as-Judge]
\begin{lstlisting}[]
Given a Question and its Golden Answer, verify whether the Predicted Answer is correct. The 
prediction is correct if it fully aligns with the meaning and key information of the Golden 
Answer. Respond with True if the prediction is corret and False otherwise. 

Question: {question}
Golden Answer: {ground truth} 
Predicted Answer: {prediction}
\end{lstlisting}
\end{tcolorbox}

\section{Ablation Studies}
\subsection{Effect of Threshold on Retriever-based Terminator}
\label{app:tau}
We conduct an ablation study to examine the effect of the threshold ($\tau$) in the retriever-based terminator. As shown in~\autoref{tab:threshold}, the threshold controls the termination criteria: A larger threshold $\tau$ makes the stopping criterion stricter, since only highly similar sub-queries will be considered redundant. This leads to later termination and thus more iterations. With $\tau=0.8$, the model halts earlier, yielding shorter inference time (\eg 6.5s on HotpotQA) but lower accuracy. In contrast, $\tau=0.9$ permits more iterations, improving EM and $ACC_L$ but at the cost of increased latency (\eg 12.7s on 2WikiMultiHop). The intermediate setting $\tau=0.85$ provides a favorable balance, delivering competitive accuracy with moderate inference time. Overall, these results confirm a trade-off: stricter termination accelerates inference but limits reasoning, while looser termination enhances accuracy at the expense of efficiency. $\tau=0.85$ emerges as a robust default across datasets.

\setlength{\tabcolsep}{6pt}
\begin{table*}[t]
  \centering
  \small
  \begin{tabular}{@{}lccccccccc@{}}
    \toprule
                  &       \multicolumn{3}{c}{HotpotQA}                            &     \multicolumn{3}{c}{2WikiMultiHop}                                  &   \multicolumn{3}{c}{MuSiQue}        \\ 
    $\tau$        &  EM($\uparrow$) &  $ACC_L$($\uparrow$) & $T$($\downarrow$) &  EM($\uparrow$) &  $ACC_L$($\uparrow$) & $T$($\downarrow$)  &  EM($\uparrow$) &  $ACC_L$($\uparrow$) & $T$($\downarrow$)  \\
    \cmidrule(lr){1-1} \cmidrule(lr){2-4} \cmidrule(lr){5-7} \cmidrule(lr){8-10}
    0.8           &    32.7    &    50.0    &    6.5    &    32.2    &    40.9    &    7.3    &    14.4    &    22.8    &    7.9     \\
    0.85          &    34.1    &    50.9    &    8.1    &    33.6    &    42.3    &    9.0    &    14.5    &    22.8    &    9.5     \\
    0.9           & 34.5         &       51.4          &     9.8       &       34.3       &      43.4       &       12.7     &     13.4        &       21.2     &     10.8 \\
    
    \bottomrule
    
    \end{tabular}
    \caption{\textbf{Effect of threshold ($\tau$) on performance and inference time across datasets.}
    A higher threshold corresponds to looser termination criteria, allowing more iterations before answering the question. This can improve EM and $ACC_L$, but also increases inference time, reflecting the trade-off between accuracy and efficiency.
    }
    
  \label{tab:threshold}
\end{table*}

\subsection{Performance vs. Efficiency across Datasets}
\label{app:ablation}
\begin{figure}[t]
    \centering
    \includegraphics[width=\linewidth]{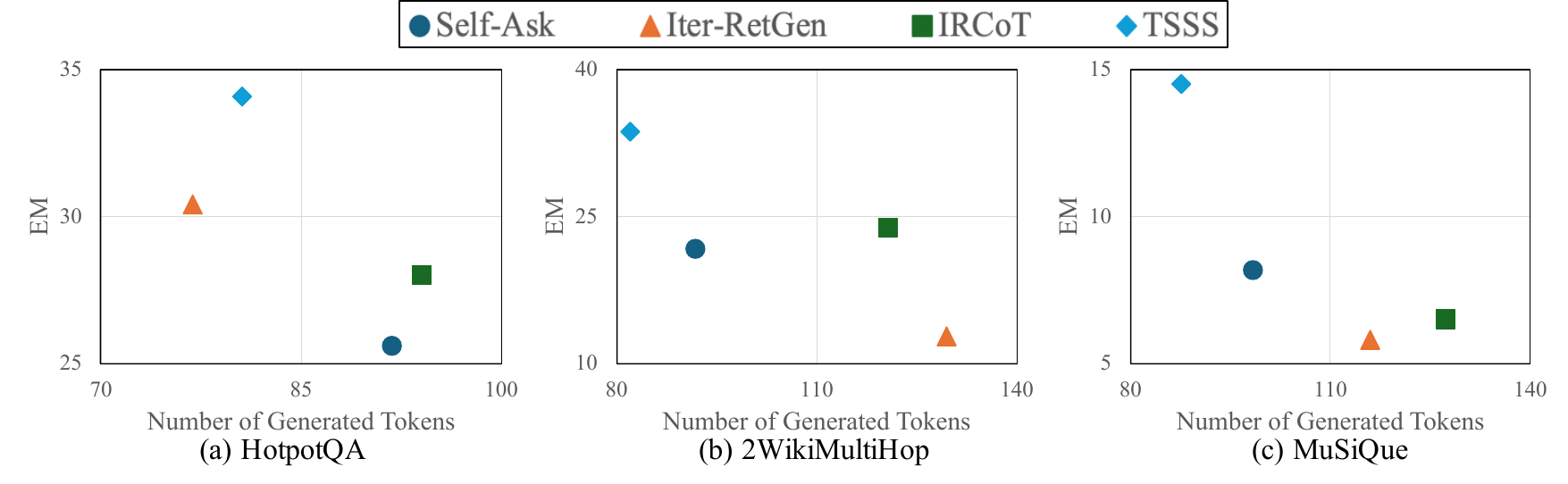}
    \vspace{-0.6cm}
    \caption{\textbf{Performance–efficiency trade-off of different methods on three datasets.}~\myalg achieves the highest performance (EM) with fewer tokens than other baselines, demonstrating a superior balance of performance and efficiency.}
    \label{fig:app_em_ng}
    \vspace{-0.5cm}
\end{figure}

\autoref{fig:app_em_ng} further examines the trade-off between accuracy and efficiency on HotpotQA, 2WikiMultiHop, and MuSiQue. On HotpotQA, \myalg achieves the highest EM score, while Iter-RetGen generates fewer tokens but suffers a clear performance drop, showing that reducing token usage alone is insufficient without preserving strong reasoning ability. In contrast, on both 2WikiMultiHop and MuSiQue, \myalg simultaneously achieves the fewest generated tokens and the highest EM among all baselines, highlighting its superior balance of performance and efficiency in more complex multi-hop settings. Taken together, these results confirm that \myalg consistently delivers state-of-the-art accuracy while maintaining efficiency, offering the most favorable trade-off under resource-constrained settings.

\section{Limitations \& Future Work}
\label{app:futurework}
While our framework alleviates key inefficiencies in multi-hop RAG, the termination mechanism still depends on heuristic redundancy detection, which may not capture all cases where additional reasoning is necessary. A promising direction is to train LLMs to self-terminate, enabling more reliable and adaptive stopping. Beyond this, an exciting avenue for future work is the development of \textit{adaptive templates} that extend the benefits of our structured reasoning beyond multi-hop QA. Such templates could dynamically adjust to diverse tasks while maintaining efficiency, offering the potential to preserve high accuracy under on-device constraints by further reducing the computational cost of reasoning. This direction highlights the broader vision of enabling lightweight yet powerful reasoning systems deployable across a wide range of real-world scenarios.

%% file: neurips_2025.bbl
\begin{thebibliography}{10}

\bibitem{grattafiori2024llama}
Aaron Grattafiori, Abhimanyu Dubey, Abhinav Jauhri, Abhinav Pandey, Abhishek Kadian, Ahmad Al-Dahle, Aiesha Letman, Akhil Mathur, Alan Schelten, Alex Vaughan, et~al.
\newblock The llama 3 herd of models.
\newblock {\em ArXiv Preprint}, 2024.

\bibitem{team2025gemma}
Gemma Team, Aishwarya Kamath, Johan Ferret, Shreya Pathak, Nino Vieillard, Ramona Merhej, Sarah Perrin, Tatiana Matejovicova, Alexandre Ram{\'e}, Morgane Rivi{\`e}re, et~al.
\newblock Gemma 3 technical report.
\newblock {\em ArXiv Preprint}, 2025.

\bibitem{abdin2024phi}
Marah Abdin, Jyoti Aneja, Harkirat Behl, S{\'e}bastien Bubeck, Ronen Eldan, Suriya Gunasekar, Michael Harrison, Russell~J Hewett, Mojan Javaheripi, Piero Kauffmann, et~al.
\newblock Phi-4 technical report.
\newblock {\em ArXiv Preprint}, 2024.

\bibitem{bang2024crayon}
Jihwan Bang, Juntae Lee, Kyuhong Shim, Seunghan Yang, and Simyung Chang.
\newblock Crayon: Customized on-device llm via instant adapter blending and edge-server hybrid inference.
\newblock In {\em Proceedings of the Association for Computational Linguistics}, pages 3720--3731, 2024.

\bibitem{dang2025reinforcement}
Quy-Anh Dang and Chris Ngo.
\newblock Reinforcement learning for reasoning in small llms: What works and what doesn't.
\newblock {\em ArXiv Preprint}, 2025.

\bibitem{lee2025chain}
Juntae Lee, Jihwan Bang, Kyuhong Shim, Seunghan Yang, and Simyung Chang.
\newblock Chain-of-rank: Enhancing large language models for domain-specific rag in edge device.
\newblock In {\em Proceedings of the Nations of the Americas Chapter of the Association for Computational Linguistics}, pages 5601--5608, 2025.

\bibitem{yang2018HotpotQA}
Zhilin Yang, Peng Qi, Saizheng Zhang, Yoshua Bengio, William Cohen, Ruslan Salakhutdinov, and Christopher~D Manning.
\newblock Hotpotqa: A dataset for diverse, explainable multi-hop question answering.
\newblock In {\em Proceedings of Empirical Methods in Natural Language Processing}, pages 2369--2380, 2018.

\bibitem{trivedi2022MuSiQue}
Harsh Trivedi, Niranjan Balasubramanian, Tushar Khot, and Ashish Sabharwal.
\newblock ♫ musique: Multihop questions via single-hop question composition.
\newblock {\em Transactions of the Association for Computational Linguistics}, 10:539--554, 2022.

\bibitem{xanh2020_2wikimultihop}
Xanh Ho, Anh-Khoa Duong~Nguyen, Saku Sugawara, and Akiko Aizawa.
\newblock Constructing a multi-hop {QA} dataset for comprehensive evaluation of reasoning steps.
\newblock In {\em Proceedings of International Conference on Computational Linguistics}, pages 6609--6625, 2020.

\bibitem{shao2023enhancing}
Zhihong Shao, Yeyun Gong, Yelong Shen, Minlie Huang, Nan Duan, and Weizhu Chen.
\newblock Enhancing retrieval-augmented large language models with iterative retrieval-generation synergy.
\newblock In {\em Proceedings of Empirical Methods in Natural Language Processing (Findings)}, pages 9248--9274, 2023.

\bibitem{trivedi2023interleaving}
Harsh Trivedi, Niranjan Balasubramanian, Tushar Khot, and Ashish Sabharwal.
\newblock Interleaving retrieval with chain-of-thought reasoning for knowledge-intensive multi-step questions.
\newblock In {\em Proceedings of the Association for Computational Linguistics}, pages 10014--10037, 2023.

\bibitem{press2023measuring}
Ofir Press, Muru Zhang, Sewon Min, Ludwig Schmidt, Noah~A Smith, and Mike Lewis.
\newblock Measuring and narrowing the compositionality gap in language models.
\newblock In {\em Proceedings of Empirical Methods in Natural Language Processing}, pages 5687--5711, 2023.

\bibitem{wu2025composerag}
Ruofan Wu, Youngwon Lee, Fan Shu, Danmei Xu, Seung-won Hwang, Zhewei Yao, Yuxiong He, and Feng Yan.
\newblock Composerag: A modular and composable rag for corpus-grounded multi-hop question answering.
\newblock {\em ArXiv Preprint}, 2025.

\bibitem{jiang2025retrieve}
Zhouyu Jiang, Mengshu Sun, Lei Liang, and Zhiqiang Zhang.
\newblock Retrieve, summarize, plan: Advancing multi-hop question answering with an iterative approach.
\newblock In {\em Companion Proceedings of the ACM on Web Conference 2025}, pages 1677--1686, 2025.

\bibitem{zhuang2024efficientrag}
Ziyuan Zhuang, Zhiyang Zhang, Sitao Cheng, Fangkai Yang, Jia Liu, Shujian Huang, Qingwei Lin, Saravan Rajmohan, Dongmei Zhang, and Qi~Zhang.
\newblock Efficientrag: Efficient retriever for multi-hop question answering.
\newblock In {\em Proceedings of Empirical Methods in Natural Language Processing}, pages 3392--3411, 2024.

\bibitem{douze2024faiss}
Matthijs Douze, Alexandr Guzhva, Chengqi Deng, Jeff Johnson, Gergely Szilvasy, Pierre-Emmanuel Mazaré, Maria Lomeli, Lucas Hosseini, and Hervé Jégou.
\newblock The faiss library.
\newblock 2024.

\bibitem{wang2022text}
Liang Wang, Nan Yang, Xiaolong Huang, Binxing Jiao, Linjun Yang, Daxin Jiang, Rangan Majumder, and Furu Wei.
\newblock Text embeddings by weakly-supervised contrastive pre-training.
\newblock {\em ArXiv Preprint}, 2022.

\bibitem{jin2025flashrag}
Jiajie Jin, Yutao Zhu, Zhicheng Dou, Guanting Dong, Xinyu Yang, Chenghao Zhang, Tong Zhao, Zhao Yang, and Ji-Rong Wen.
\newblock Flashrag: A modular toolkit for efficient retrieval-augmented generation research.
\newblock In {\em Companion Proceedings of the ACM on Web Conference}, pages 737--740, 2025.

\bibitem{kimsure}
Jaehyung Kim, Jaehyun Nam, Sangwoo Mo, Jongjin Park, Sang-Woo Lee, Minjoon Seo, Jung-Woo Ha, and Jinwoo Shin.
\newblock Sure: Summarizing retrievals using answer candidates for open-domain qa of llms.
\newblock In {\em Proceedings of International Conference on Learning Representations}, 2024.

\bibitem{xu2023recomp}
Fangyuan Xu, Weijia Shi, and Eunsol Choi.
\newblock Recomp: Improving retrieval-augmented lms with compression and selective augmentation.
\newblock {\em Proceedings of International Conference on Learning Representations}, 2024.

\bibitem{shi2024replug}
Weijia Shi, Sewon Min, Michihiro Yasunaga, Minjoon Seo, Richard James, Mike Lewis, Luke Zettlemoyer, and Wen-tau Yih.
\newblock Replug: Retrieval-augmented black-box language models.
\newblock In {\em Proceedings of the Conference of the North American Chapter of the Association for Computational Linguistics: Human Language Technologies (Volume 1: Long Papers)}, pages 8364--8377, 2024.

\bibitem{zheng2023judging}
Lianmin Zheng, Wei-Lin Chiang, Ying Sheng, Siyuan Zhuang, Zhanghao Wu, Yonghao Zhuang, Zi~Lin, Zhuohan Li, Dacheng Li, Eric Xing, et~al.
\newblock Judging llm-as-a-judge with mt-bench and chatbot arena.
\newblock In {\em Proceedings of Advances in Neural Information Processing Systems}, volume~36, pages 46595--46623, 2023.

\bibitem{hurst2024gpt}
Aaron Hurst, Adam Lerer, Adam~P Goucher, Adam Perelman, Aditya Ramesh, Aidan Clark, AJ~Ostrow, Akila Welihinda, Alan Hayes, Alec Radford, et~al.
\newblock Gpt-4o system card.
\newblock {\em ArXiv Preprint}, 2024.

\end{thebibliography}
